\documentclass[letterpaper, 10 pt, conference]{ieeeconf}  

\IEEEoverridecommandlockouts                              

\usepackage{graphics} 
\usepackage{epsfig} 
\usepackage{times} 
\usepackage{amsmath} 
\usepackage{amssymb}  
\usepackage{multirow}
\usepackage{float}
\usepackage{booktabs}
\usepackage{makecell}
\usepackage[hidelinks]{hyperref}

\title{\LARGE \bf
NaviDriveVLM: Decoupling High-Level Reasoning and Motion Planning for Autonomous Driving
}

\author{Ximeng Tao$^{1}$, Pardis Taghavi$^{1}$, Dimitar Filev$^{1}$, Reza Langari$^{1}$, Gaurav Pandey$^{2}$
\thanks{$^{1}$Ximeng Tao, Pardis Taghavi, Dimitar Filev and Reza Langari are with J. Mike Walker '66 Department of Mechanical Engineering, Texas A\&M University, College Station, TX 77843, USA
        {\tt\small ximeng, ptgh, dfilev, rlangari@tamu.edu}}%
\thanks{$^{2}$Gaurav Pandey is with The Department of Engineering Technology and Industrial Distribution Texas A\&M University, College Station, TX 77843, USA
        {\tt\small gpandey@tamu.edu}}%
}

\begin{document}

\maketitle
\thispagestyle{empty}
\pagestyle{empty}

\begin{abstract}

Vision-language models~(VLMs) have emerged as a promising direction for end-to-end autonomous driving~(AD) by jointly modeling visual observations, driving context, and language-based reasoning. However, existing VLM-based systems face a trade-off between high-level reasoning and motion planning: large models offer strong semantic understanding but are costly to adapt for precise control, whereas small VLM models can be fine-tuned efficiently but often exhibit weaker reasoning. We propose NaviDriveVLM, a decoupled framework that separates reasoning from action generation using a large-scale Navigator and a lightweight trainable Driver. This design preserves reasoning ability, reduces training cost, and provides an explicit interpretable intermediate representation for downstream planning. Experiments on the nuScenes benchmark show that NaviDriveVLM outperforms large VLM baselines in end-to-end motion planning. 
Codes are available at: \url{https://github.com/TAMU-CVRL/NaviDrive}.

\end{abstract}

\section{INTRODUCTION}
\begin{figure}[h!]
    \centering
    \includegraphics[width=0.95\linewidth]{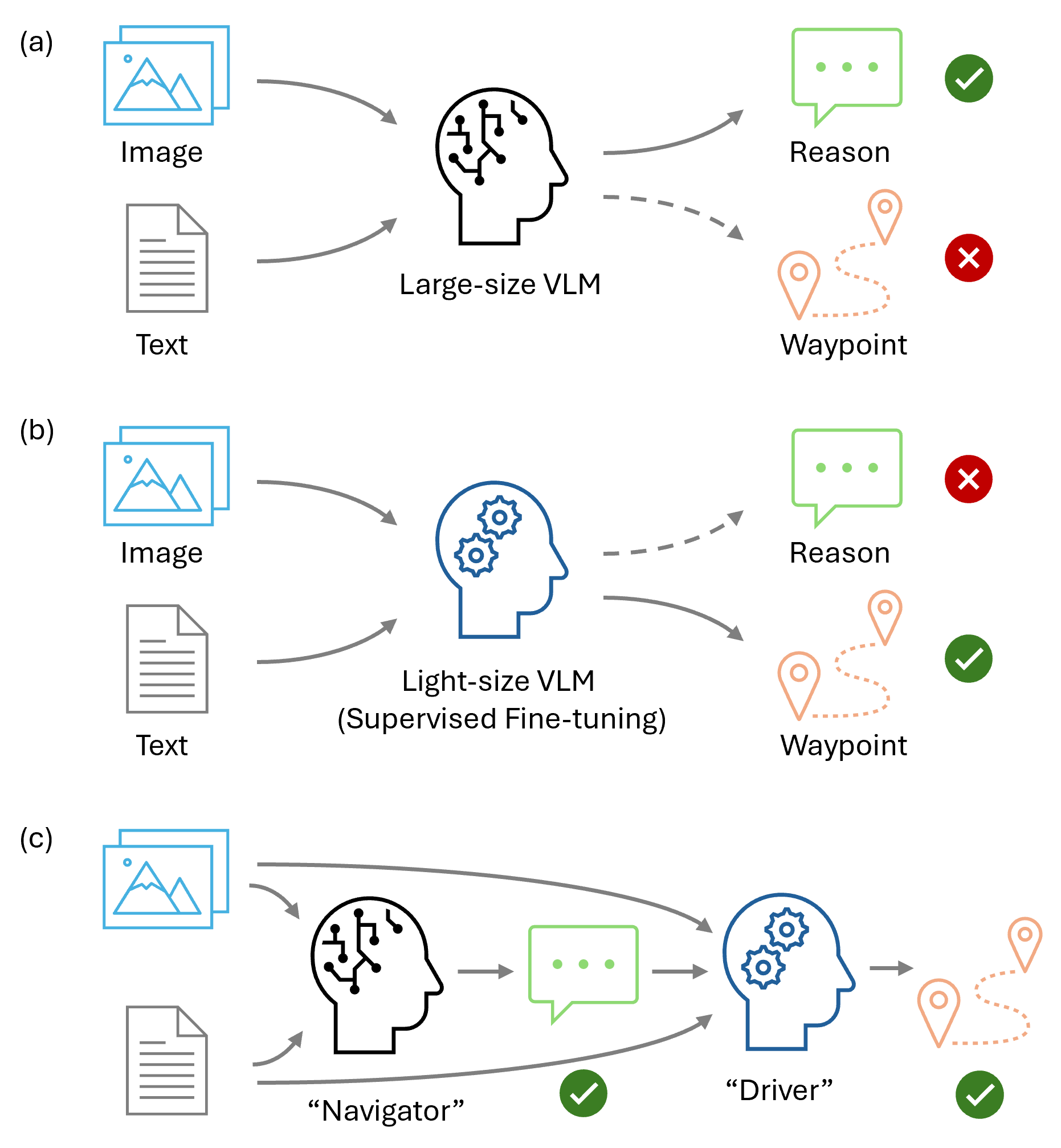}
    \caption{(a) Large-scale VLMs show strong reasoning ability but fail to generate accurate driving actions without fine-tuning. (b) Lightweight VLMs can be fine-tuned for future waypoints prediction, but their reasoning ability degrades. (c) We decouple reasoning and motion planning into two modules, using a large-scale VLM as the Navigator for reasoning and a lightweight VLM as the Driver for waypoint prediction, preserving reasoning while optimizing driving performance.}
    \label{Intro}
\end{figure}

End-to-end autonomous driving (AD) has evolved from systems focused primarily on perception~\cite{taghavi2024swinmtl} to frameworks that must jointly understand scene context, follow navigation intent, and generate reliable future actions~\cite{hu2023planning, li2024hydra}. In parallel, recent advances that incorporate language have opened a promising direction for AD by combining visual perception with semantic reasoning and language-guided decision making. Early efforts explored language-based decision making for driving~\cite{mao2023language}, while subsequent vision-language and vision-language-action models further unified perception, reasoning, and trajectory generation within a single architecture~\cite{chen2024driving, huang2024drivegpt, tian2024drivevlm, hwang2024emma, zhou2025opendrivevla, yang2025drivemoe}. More recent systems continue to strengthen this trend through improved reasoning quality, action generation, and multimodal specialization~\cite{wang2025alpamayo, ma2025leapvad, xu2025drivegpt4}. Compared with conventional modular systems, these models offer the appealing possibility of improved interpretability, since they can explicitly describe traffic scenes, justify decisions, and expose intermediate reasoning. Such properties are particularly valuable in safety critical driving, where transparent decision making is as important as planning accuracy.

However, existing VLM-based driving systems still face a basic trade-off between high-level reasoning and motion planning. Larger models are often better at understanding the scene and semantic reasoning, but they are not naturally optimized for precise motion prediction or direct action generation without costly task-specific adaptation~\cite{hwang2024emma, zhou2025opendrivevla, yang2025drivemoe}. In contrast, smaller models can be fine-tuned more efficiently for waypoint or action prediction, but they often need extra supervision, reasoning transfer, or distillation to recover the benefits of stronger semantic guidance~\cite{xu2024vlm, hegde2025distilling, feng2025verdi, liu2025dsdrive, ma2025leapvad}. As a result, using a single model to jointly perform reasoning and control can lead to a difficult balance between reasoning quality, adaptation efficiency, and planning accuracy.

To address this challenge, we propose \textbf{NaviDriveVLM}, a decoupled framework that separates semantic reasoning from action generation. It consists of a frozen large-scale VLM, referred to as the \emph{Navigator}, and a lightweight trainable VLM, referred to as the \emph{Driver}. The Navigator takes surround view images, ego states, and task prompts as input, and produces semantic guidance in the form of a scene description, a recommended action, and the corresponding reasoning. The Driver then uses this reasoning output, images, ego states, and task prompts to predict future waypoints. Keeping the Navigator frozen preserves strong reasoning ability while avoiding the computational burden of retraining a large model. The specialized Driver can be adapted efficiently for downstream motion prediction. This
design turns semantic reasoning into an explicit and interpretable intermediate representation between perception and planning.

We evaluate NaviDriveVLM for end-to-end motion planning task on the nuScenes benchmark~\cite{nuscenes2019}. Our results show that the decoupled design outperforms single large-VLM baselines that directly fine-tune a standalone model for trajectory prediction, and our ablations further show that the Navigator’s reasoning improves planning performance. Together, these findings indicate that separating semantic reasoning from motion planning is a useful design choice for building VLM-based AD systems that are both more interpretable and more effective.

Our contributions are threefold: 1) we introduce NaviDriveVLM, a decoupled Navigator--Driver framework for autonomous driving; 2) we show that structured reasoning can serve as an interpretable intermediate representation for improving waypoint prediction; and 3) we evaluate NaviDriveVLM on the nuScenes benchmark and show that the decoupled design achieves stronger planning performance than single large-VLM baselines, while retaining interpretability and reducing adaptation cost.

\section{RELATED WORK}

\subsection{VLMs for End-to-End Autonomous Driving}

Recent end-to-end driving research has increasingly incorporated language modeling to improve generalization and semantic understanding. Early planning-oriented frameworks such as UniAD~\cite{hu2023planning} unified perception, prediction, and planning toward the final driving objective, while language-driven systems such as GPT-Driver~\cite{mao2023gpt}, DriveMLM~\cite{wang2023drivemlm}, and LMDrive~\cite{shao2024lmdrive} showed that LLMs can support motion planning, behavior planning, and instructioncfollowing driving. More recent multimodal systems, including Driving with LLMs~\cite{chen2024driving}, DriveGPT4~\cite{huang2024drivegpt}, EMMA~\cite{hwang2024emma}, and DriveMoE~\cite{yang2025drivemoe}, further integrate visual perception, language reasoning, and trajectory prediction within a single model. Collectively, these works show the promise of VLMs for AD, but most still rely on a unified model to perform both high-level reasoning and future waypoints generation, making it difficult to simultaneously preserve strong reasoning ability, efficient adaptation, and precise control. 

\subsection{Semantic Reasoning and Explainability}

A parallel line of work uses language to improve interpretability and make driving decisions more transparent. DriveLM~\cite{sima2024drivelm} explicitly decomposes reasoning across perception, prediction, and planning. Reason2Drive~\cite{nie2024reason2drive} advances this direction with a large-scale benchmark for interpretable, chain-based reasoning in driving scenes, while RAG-Driver~\cite{yuan2024rag} improves explanation quality through retrieval augmented reasoning. Related systems such as Alpamayo~\cite{wang2025alpamayo} and DriveGPT4-v2~\cite{xu2025drivegpt4} also generate natural-language rationales together with actions or decisions. These studies establish language is valuable not only for supervision but also for exposing intermediate decision logic; however, in most prior work, reasoning primarily serves as an auxiliary explanation signal rather than a separated intermediate representation for downstream control. 

\subsection{Decoupled and Hierarchical Driving Architectures}

Separating high-level decision-making from motion planning is a long-standing principle in autonomous driving, and several recent driving systems adopt modular forms of integration between reasoning and planning. Hydra-MDP~\cite{li2024hydra} uses a teacher-student formulation with multi-target hydra distillation to learn diverse trajectory candidates, while VLM-AD~\cite{xu2024vlm} and DiMA~\cite{hegde2025distilling} leverage VLM/MLLM supervision during training to improve a lighter planner. More recent methods such as DSDrive~\cite{liu2025dsdrive} and LeapVAD~\cite{ma2025leapvad} further emphasize integration of reasoning and planning through distillation or dual-process design. While recent modular architectures often utilize reasoning as a supervision signal during training, NaviDriveVLM employs reasoning as an explicit and interpretable intermediate representation. By decoupling a frozen, large-scale Navigator from a lightweight, trainable Driver, we preserve complex semantic guidance while achieving precise motion planning.


\section{METHODOLOGY}
\begin{figure*}
    \centering
    \vspace{4pt}
    \includegraphics[width=1\linewidth]{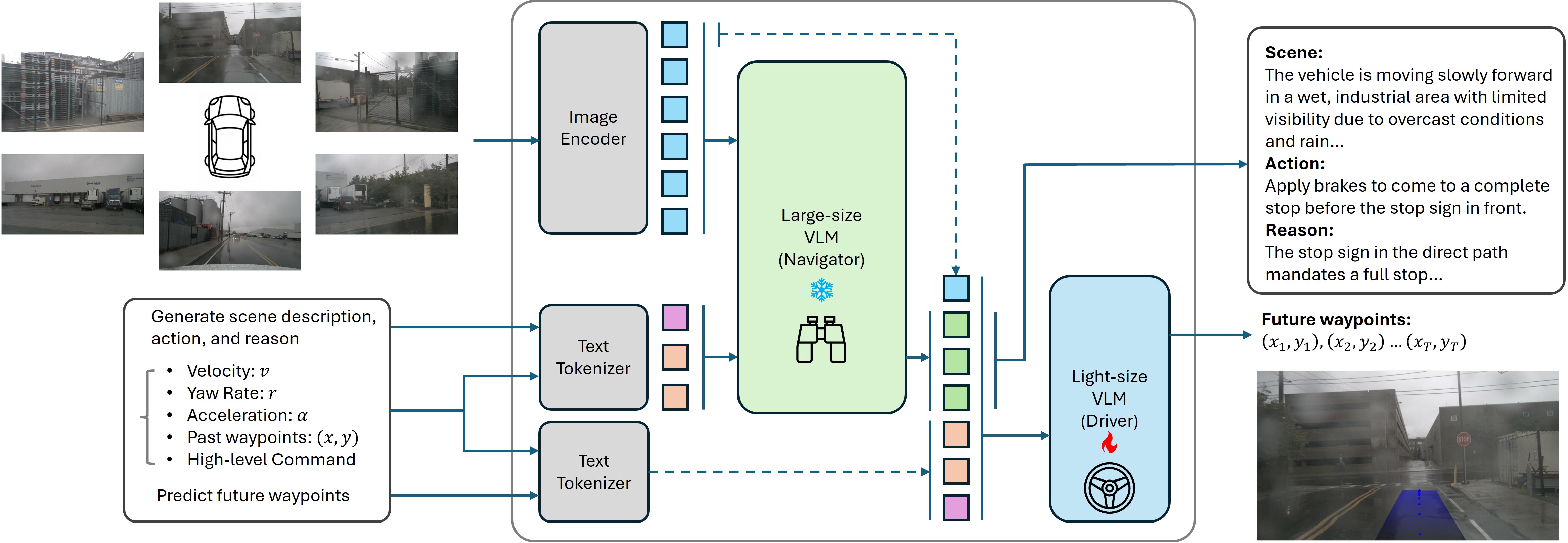}
    \caption{Overview of NaviDriveVLM. The framework is decoupled into a large VLM serving as the Navigator and a lightweight VLM serving as the Driver. (a) Multi-view surround images are encoded into visual tokens (blue). The Navigator prompt and ego state are tokenized as text tokens (pink and orange). (b) The Navigator VLM generates reasoning tokens (green), which are concatenated with the front-view image tokens, Driver prompt, and ego state tokens, and then fed into the Driver VLM. (c) The Driver VLM is fine-tuned to predict future waypoints or driving actions. The reasoning tokens can be decoded into text for interpretability.}
    \label{Pipeline}
\end{figure*}
The overall architecture of our framework is shown in Fig.~\ref{Pipeline}. The system is decoupled into two modules, the \emph{Navigator} and the \emph{Driver}. The Navigator is a large-scale VLM responsible for scene understanding and high-level reasoning. In principle, it can be replaced by any advanced reasoning-capable VLM.
To preserve its inherent reasoning capability and avoid the substantial computational cost associated with retraining large-scale parameters, the Navigator remains frozen during training. 
The Driver is a lightweight VLM, which enables efficient fully supervised fine-tuning (SFT) as a driving expert for future waypoint prediction.

\subsection{Navigator}

A pretrained large scale VLM is used as a \textit{Navigator} which generates the scene description, action and reasoning for the suggested action. It takes the multi-view surrounding images ($\mathcal{I}$), the ego-vehicle state ($O_{ego}$ = [velocity ($v$), yaw rate ($r$), acceleration ($\alpha$)]), past waypoints $(x_t,y_t)$, and a high-level command as input (Fig.~\ref{Pipeline}).
The high-level command is determined based on the ground-truth waypoints and recorded in the training dataset. 
It categorizes driver intention into six classes: \texttt{Hard Left, Slight Left, Keep Straight, Slight Right, Hard Right, and Decelerate Stop}.
The user prompt and system prompt for this Navigator VLM (Navi-VLM) are represented as $Q_{N}$ and $S_{N}$, respectively, with detailed content shown in the upper part of Fig.~\ref{Prompt}.
The Navi-VLM is denoted as $\mathcal{G}_{N}$, and the reasoning generation process can be described as:
\begin{equation}
O_{R} = \mathcal{G}_{N}(\mathcal{I}, O_{ego}, Q_N, S_N).
\label{Navi}
\end{equation}
The reasoning output $O_{R}$ consists of three components: scene description, recommended action, and corresponding reasoning explanation.
These reasoning tokens can be directly forwarded to the lightweight Driver VLM (Driver-VLM) to assist future waypoint prediction and improve planning performance, or decoded to obtain human-readable explanations.

\subsection{Driver}
\begin{figure}[!htbp]
    \centering
    \vspace{4pt}
    \includegraphics[width=0.95\linewidth]{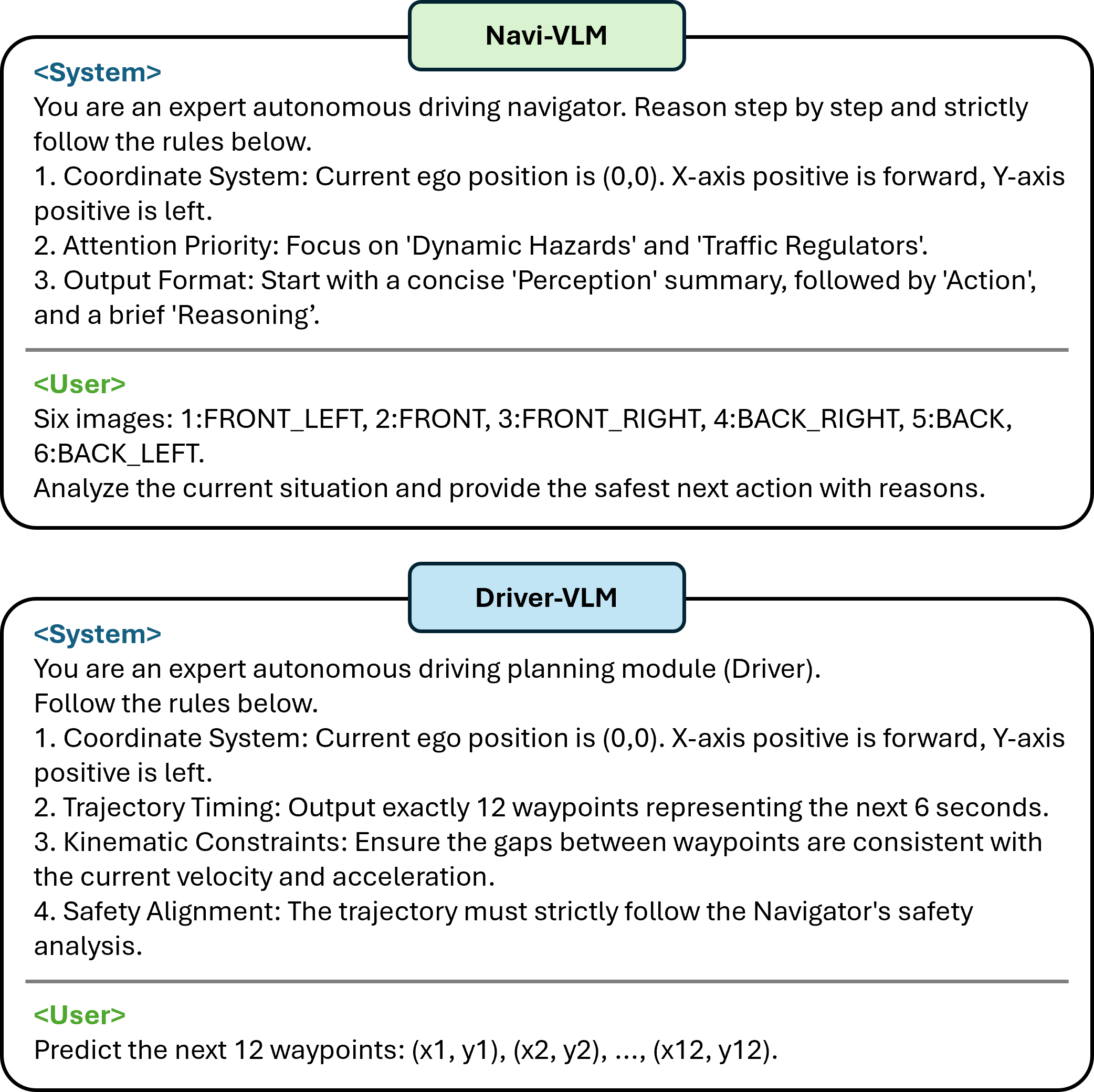}
    \caption{Prompt design for the Navi-VLM and Driver-VLM.}
    \label{Prompt}
\end{figure}

A lightweight VLM is fully fine-tuned to serve as a driver expert, enabling it to understand reasoning outputs and observational inputs to predict future waypoints ($\mathcal{W} = \{w_t = (x_t, y_t)\}_{t=1}^{T}$).
In Eq.~\ref{Driver}, in addition to the four common inputs $(\mathcal{I}, O_{ego}, Q_D, S_D)$, the reasoning output $O_{R}$ is introduced.
The system and user prompts differ from those used in Navi-VLM and are shown in the lower part of Fig.~\ref{Prompt}.
The reasoning output $O_{R}$ generated by the Navi-VLM is incorporated as an auxiliary input. 
The reasoning information plays an important role in improving prediction accuracy.
\begin{equation}
\mathcal{W} = \mathcal{G}_{D}(O_{R}, \mathcal{I}, O_{ego}, Q_D, S_D).
\label{Driver}
\end{equation}
The VLM-based approach models waypoint prediction as a probabilistic generation process, formulated as
\begin{equation}
P(\mathcal{W} \mid O_{R}, \mathcal{I}, O_{ego}, Q_D, S_D),
\end{equation}
where $\mathcal{W}$ denotes the future waypoint sequence.
During the SFT stage, we optimize the Driver-VLM parameters $\theta_D$ by minimizing the negative log-likelihood of the ground-truth waypoint sequence $\mathcal{W}$. $w_t$ is the target waypoint to be predicted at frame $t$, and $w_{<t}$ represent the ground-truth waypoints from previous time steps. The loss function is defined as:
\begin{equation}
\mathcal{L}_{SFT} =
-\sum_{t=1}^{T}
\log P(w_t \mid w_{<t}, O_{R}, \mathcal{I}, O_{ego}, Q_D, S_D; \theta_D).
\end{equation}
The ground-truth waypoints are used as assistant prompts during training, and non-target tokens are masked under the autoregressive training objective.

\section{EXPERIMENTS}
\label{experiment}
\begin{figure*}[!htbp]
    \centering
    \vspace{4pt}
    \includegraphics[width=1\linewidth]{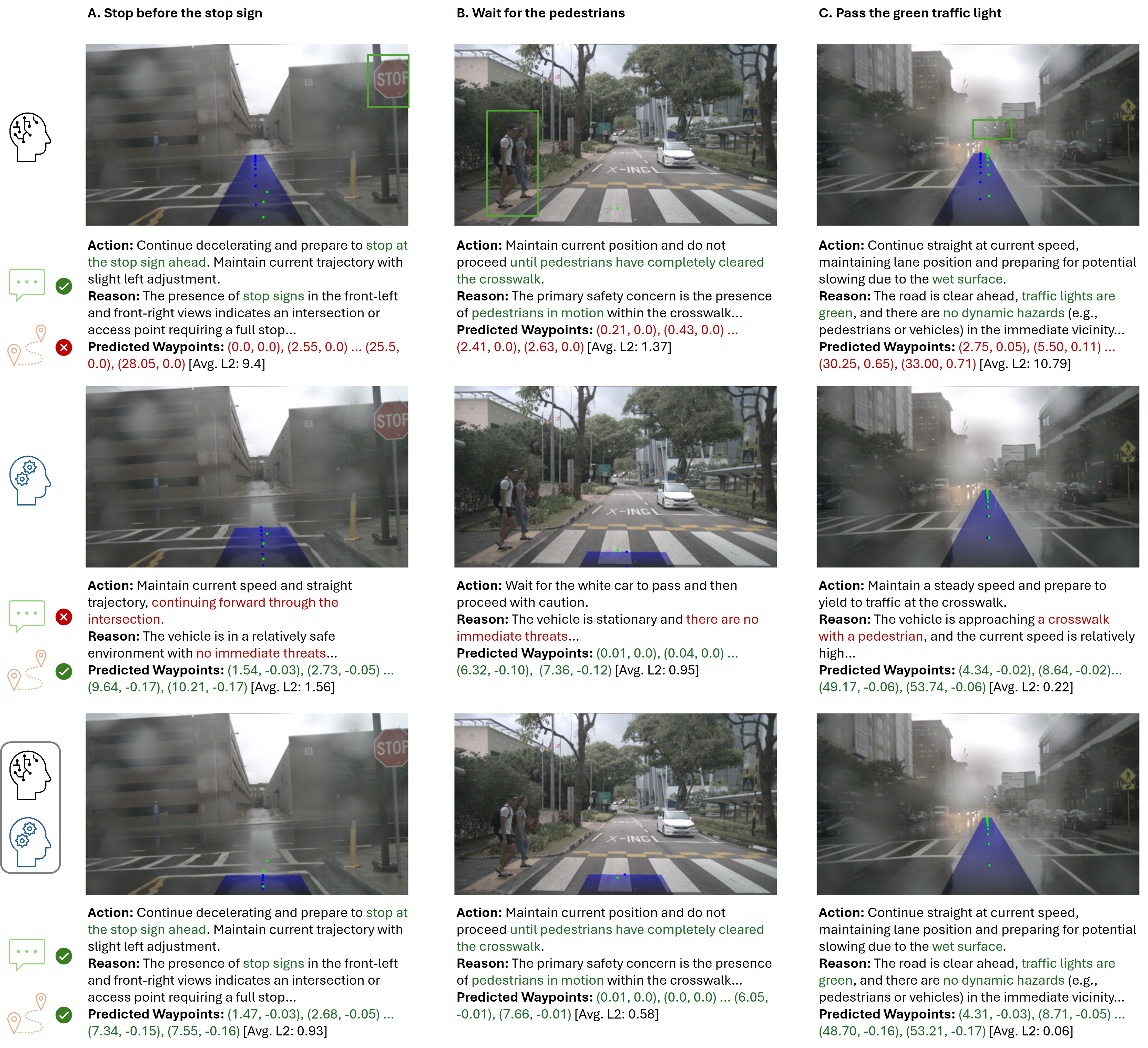}
    \caption{Qualitative results across three different driving scenarios.
    Predicted waypoints are visualized as blue masks and blue dots, while ground-truth waypoints are shown in green.
    The minimum average L2 error in meters over 6 seconds is shown in brackets.
    The first row represents a non-fine-tuned large VLM, which is capable of generating reasonable high-level reasoning outputs. However, the predicted future waypoints deviate significantly from the ground truth.
    The second row corresponds to a smaller fine-tuned VLM, which can generate accurate future waypoints suitable for control. However, it lacks strong scene understanding and reasoning capabilities.
    The third row presents our proposed NaviDriveVLM framework, which combines reliable high-level reasoning with accurate future waypoint prediction.
    }
    \label{Results}
\end{figure*}

\label{experiment}
\begin{figure*}[!htbp]
    \centering
    \vspace{4pt}
    \includegraphics[width=1\linewidth]{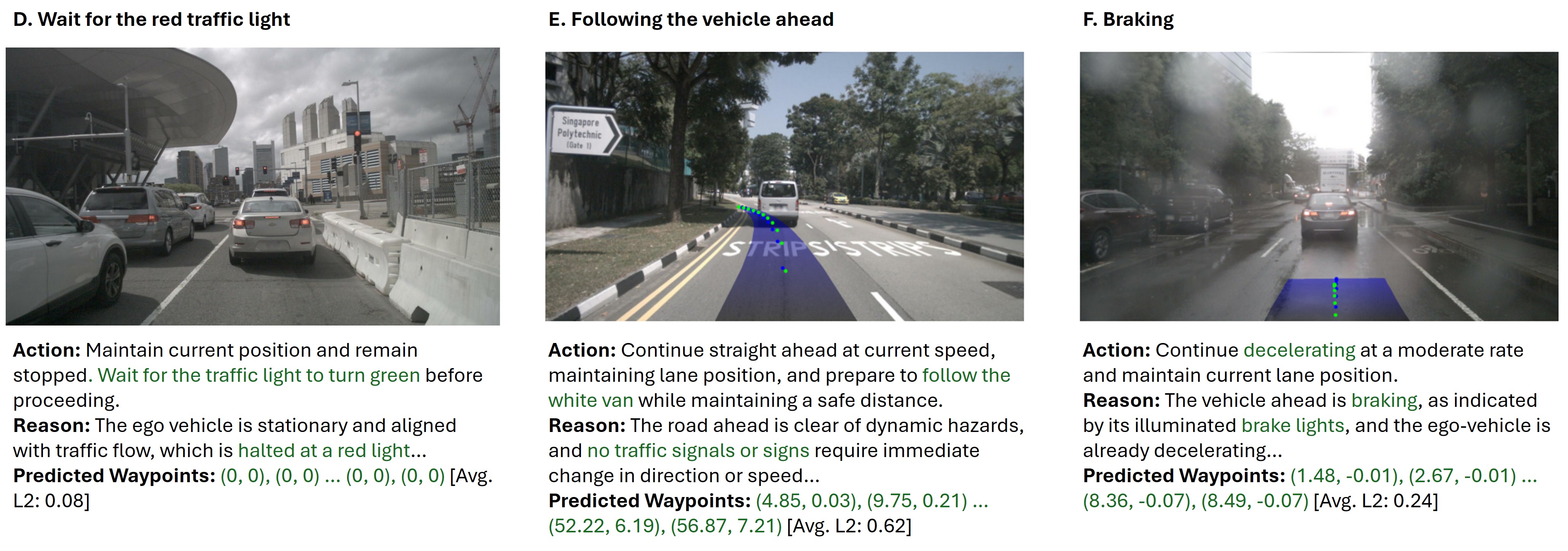}
    \caption{Additional qualitative results from NaviDriveVLM.
    Predicted waypoints are visualized as blue masks and blue dots, while ground-truth waypoints are shown in green.
    The minimum average L2 error in meters over 6 seconds is shown in brackets.
    Scenes D, E, and F illustrate the reasoning results and the corresponding predicted future waypoints predicted by NaviDriveVLM across three scenarios: waiting at a red traffic light, following another vehicle, and braking.
    }
    \label{More_Results}
\end{figure*}

\subsection{Dataset}
We use the nuScenes dataset~\cite{nuscenes2019} to construct a derived dataset, referred to as the nuScenes-Reason dataset.
The nuScenes contains 850 scenes, where each scene consists of 20-second driving sequences sampled at 2 Hz. 
We employ an 8-second sliding window to extract thirteen 8-second clips from each scene, which are further divided into a 2-second historical context and a 6-second future prediction horizon. 
Meanwhile, the waypoints, heading, velocity, yaw rate, acceleration, and images associated with these clips are recorded for further training and evaluation.
This procedure yields 16.54k training samples and 3618 test samples.
To accelerate the training process, we decouple training into two stages.
First, a large pre-trained VLM (NaviVLM) is used to generate reasoning outputs for each 8-second clip. The generated reasoning results are recorded as text, constructing the nuScenes-Reason dataset.
Second, Driver-VLM is trained using the nuScenes-Reason dataset, avoiding repeated Navi-VLM inference and improving training efficiency.

\begin{table*}[!htbp]
\centering
\caption{End-to-end motion planning experiments on nuScenes}
\label{tab:results}
\scalebox{1.0}{
\begin{tabular}{l|c|ccc|c}
\toprule
Model & VLM & L2 (1s) & L2 (2s) & L2 (3s) & Avg. L2 (m) $\downarrow$ \\
\midrule
OpenEMMA~\cite{xing2025openemma} & LLaVa-7B~\cite{liu2023visual}  & 1.45 & 3.21 & 3.76 & 2.81\\
ST-P3~\cite{hu2022st} & - & 1.33 & 2.11 & 2.90 & 2.11 \\
GenAD~\cite{zheng2024genad} & - & 0.36 &0.83 &1.55 & 0.91\\

Ego-MLP~\cite{zhai2023rethinking} & - & 0.46 & 0.76 & 1.12 & 0.78\\
VAD-Base~\cite{jiang2023vad} & - & 0.41 & 0.70& 1.05 & 0.72 \\
UniAD~\cite{hu2023planning} & - & 0.44 & 0.67 & 0.96 & 0.69 \\
Verdi~\cite{feng2025verdi} & - & 0.36 & 0.62& 0.96 & 0.65\\
\midrule
Driver-VLM & Qwen3-VL-8B & 0.24 & 0.65 & 1.25 & 0.60\\
\multirow{1}{*}{NaviDriveVLM} & Qwen3-VL-2B & \textbf{0.20} & \textbf{0.50} & \textbf{0.93} & \textbf{0.46}\\
\bottomrule
\end{tabular}}
\end{table*}

\subsection{Training}
Qwen3~\cite{qwen3technicalreport} performs well among open-source VLMs and is therefore adopted as the backbone model for both Navi-VLM and Driver-VLM. During training, only the Driver-VLM is fine-tuned using SFT.
We employ the AdamW optimizer~\cite{loshchilov2017decoupled} with a weight decay of 0.01 and an initial learning rate of $1\times10^{-5}$. A cosine learning rate schedule is adopted. The batch size is set to 1, and the model is trained for 3 epochs. Gradient accumulation is performed over 16 steps.
For the 8B-scale model, 8-bit quantization and LoRA adaptation~\cite{hu2022lora} are applied, with a LoRA rank of 64, LoRA alpha of 128, and LoRA dropout rate of 0.05. All experiments are conducted on a single NVIDIA RTX 4090 GPU.

\subsection{Metrics}
For Navi-VLM, evaluating the correctness of reasoning outputs is inherently subjective and difficult to quantify. Therefore, we primarily rely on the general reasoning capability of modern large-scale VLMs and provide qualitative analysis of representative results in Section~\ref{sec:Results}.
For the overall NaviDriveVLM framework, we focus on the open-loop motion planning task. The model generates six candidate future waypoints, and the prediction with the smallest minimum Average L2 Error at the 6-second prediction horizon is selected for evaluation.

\subsection{Results}
\label{sec:Results}

\begin{table*}[!htbp]
\centering
\vspace{4pt}
\caption{Comparison of Waypoint and Action Prediction Performance}
\label{tab:contrl}
\scalebox{1.0}{
\begin{tabular}{l|c|cccc|c}
\toprule
Model & Output & L2 (1s) & L2 (2s) & L2 (3s) & L2 (6s) & Avg. L2 (m) $\downarrow$ \\
\midrule
\multirow{2}{*}{\makecell{NaviDriveVLM}} & Waypoint $(x, y)$ & \textbf{0.200} & \textbf{0.495} & \textbf{0.934} & 3.245 & 1.285 \\
 & Action $(\alpha, \kappa)$ & 0.259 & 0.571 & 1.007 & \textbf{2.911} & \textbf{1.201}\\
\bottomrule
\end{tabular}}
\end{table*}

\begin{table*}[]
\centering
\caption{Ablation Study on Input Components of NaviDriveVLM}
\label{tab:ablation}
\scalebox{1.0}{
\begin{tabular}{l|ccc|cccc|c}
\toprule
Model & Reason & Command & Images & L2 (1s) & L2 (2s) & L2 (3s) & L2 (6s) & Avg. L2 (m) $\downarrow$ \\
\midrule
\multirow{4}{*}{{\makecell{NaviDriveVLM}}} 
 & \checkmark & & & 0.204 & 0.518 & 1.029 & 3.977 & 1.515\\
& \checkmark & & \checkmark & 0.200 & 0.516 & 1.012 & 3.861 & 1.476\\
 & \checkmark & \checkmark & & 0.200 & 0.496 & 0.934 & 3.254 & 1.288\\
 & \checkmark & \checkmark & \checkmark & \textbf{0.200} & \textbf{0.495} & \textbf{0.934} & \textbf{3.245} & \textbf{1.285}\\
\bottomrule
\end{tabular}}
\end{table*}

Fig.~\ref{Results} presents qualitative comparisons across three representative driving scenarios: A) stopping before a stop sign, B) yielding to pedestrians, and C) proceeding through a green traffic light. The first row shows results from a large VLM without supervised fine-tuning (Qwen3-VL-8B). Although this model produces reasonable semantic reasoning, it predicts inaccurate future waypoints, as reflected by the large average L2 error highlighted in red (First row, Fig.~\ref{Results}). These examples indicate that a pretrained large VLM can recognize important scene elements, such as stop signs, pedestrians in the crosswalk, and traffic light states, but without task-specific adaptation it does not generate reliable motion predictions. The second row shows results from a smaller VLM (Qwen3-VL-2B) after supervised fine-tuning. In this case, waypoint prediction becomes more accurate, but the quality of the reasoning degrades. The incorrect or incomplete reasoning is highlighted in red in the second row of Fig.~\ref{Results}. This suggests that fine-tuning a lightweight model improves prediction, but may weaken its high-level semantic reasoning. The third row presents results from the proposed NaviDriveVLM framework. By decoupling reasoning and control, NaviDriveVLM combines the strong semantic reasoning of a large VLM with the accurate waypoint prediction of a fine-tuned lightweight VLM. As shown in Fig.~\ref{Results}, this design produces both reliable reasoning outputs and trajectories that more closely match the ground truth. Fig.~\ref{More_Results} presents additional qualitative results of our proposed NaviDriveVLM framework across diverse driving scenarios.

In Table~\ref{tab:results}, we report open-loop motion planning results on the nuScenes dataset. 
At the 3-second horizon (Table~\ref{tab:results}), NaviDriveVLM outperforms several representative prior methods, including ST-P3~\cite{hu2022st}, Ego-MLP~\cite{zhai2023rethinking}, and UniAD~\cite{hu2023planning}. 
To better isolate the effect of explicit reasoning, we include a Driver-VLM baseline that removes the Navigator and uses a single VLM backbone for trajectory prediction. Compared with OpenEMMA, supervised fine-tuning of this single-model baseline leads to a clear improvement in planning accuracy. Moreover, when the Navigator is added, the performance improves further, showing that explicit semantic guidance contributes beyond supervised fine-tuning alone. 
Overall, these results highlight the importance of reasoning information in motion planning and support the effectiveness of the proposed decoupled framework for improving prediction performance.


\subsection{Waypoints vs. Control Actions}
The NaviDriveVLM is primarily trained to predict future waypoints. Additionally, we investigate an alternative formulation in which the model directly predicts driving actions. 
Since the nuScenes dataset does not provide control actions as ground-truth labels, we utilize ground-truth waypoints and convert them into corresponding control actions to serve as training supervision. To transform the discrete waypoint sequence into a continuous and kinematically feasible control sequence $\mathbf{u}$, we employ Tikhonov-regularized least-squares optimization. The objective function is defined as
\begin{equation}
\mathbf{u}^* = \arg\min_{\mathbf{u}} 
\left\| \mathbf{A}\mathbf{u} - \mathbf{b} \right\|_2^2 
+ \lambda \left\| \mathbf{L}\mathbf{u} \right\|_2^2,
\end{equation}
where $\mathbf{u} = (\alpha, \kappa)$ denotes the control actions, with $\alpha$ and $\kappa$ representing acceleration and curvature, respectively. 
The system matrix $\mathbf{A}$ and input vector $\mathbf{b}$ are derived from the kinematic model given in ~\cite{wang2025alpamayo}.
Given the initial velocity $v_0$, and initial heading $\theta_0$, the control sequence $\mathbf{u}^*$ is estimated and formatted into the final action string to be used as ground truth during training. During inference, the generated control actions are integrated to reconstruct the corresponding waypoints.

Action-based outputs are widely adopted in vision-language-action (VLA) models~\cite{black2024pi_0} by leveraging advanced generation techniques~\cite{lipman2022flow}.
In this paper, both types of outputs are directly generated using the prediction model Driver-VLM. 
Table~\ref{tab:contrl} summarizes the performance differences between the two output formulations.
Waypoint-based prediction achieves lower short-term L2 error at the 1s, 2s, and 3s horizons. 
However, for long-term prediction, direct action prediction demonstrates superior performance in terms of overall average L2.

\subsection{Ablation Studies}
Table~\ref{tab:ablation} presents the ablation studies of our framework, focusing primarily on the impact of different input configurations.
All model variants include same reasoning inputs generated from Navi-VLM. 
From the results, we observe that incorporating high-level commands reduces the average L2 from 1.515 to 1.288 (-0.227). Since Driver-VLM is fundamentally a language model, explicitly providing intention information helps guide action prediction and improves planning performance.
However, we find that incorporating image inputs does not significantly reduce average L2, which decreases only from 1.515 to 1.476 (-0.039)~\cite{li2024ego}. Despite the large number of image tokens, many may be redundant or contain limited task-relevant information, resulting in marginal performance gains.
By combining reasoning, high-level commands, and image inputs, the final NaviDriveVLM model achieves a 6-second average L2 of 1.285, outperforming all other variants.

\section{CONCLUSIONS}

In this paper, we addressed a key challenge in VLM-based end-to-end autonomous driving: how to retain strong semantic reasoning while still achieving accurate motion planning. We introduced \textbf{NaviDriveVLM}, a decoupled Navigator--Driver framework in which large VLM models provides semantic guidance and small trainable VLM models predict future waypoints or actions. By treating reasoning as an explicit and interpretable intermediate representation between perception and planning, the proposed design preserves the reasoning strengths of large models while allowing efficient adaptation for motion prediction. Experiments on the nuScenes benchmark show that the proposed decoupled design improves end-to-end motion planning over single VLM baselines, and our ablations further show that the Navigator’s reasoning contributes to these gains. Overall, these results support the idea that separating semantic reasoning from motion planning is a practical and effective direction for building autonomous driving systems that are both more interpretable and better aligned with planning performance.







\bibliographystyle{IEEEtran}
\bibliography{ref}

\end{document}